\pdfoutput=1

\documentclass[11pt]{article}

\usepackage[final]{acl}

\usepackage{times}
\usepackage{latexsym}

\usepackage[T1]{fontenc}

\usepackage[utf8]{inputenc}

\usepackage{microtype}

\usepackage{inconsolata}

\usepackage{graphicx}

\usepackage{amssymb}

\usepackage{color}
\usepackage{colortbl}

\title{\raisebox{-1.5mm}{\includegraphics[trim=0 0 -20 0, clip, width=0.8cm]{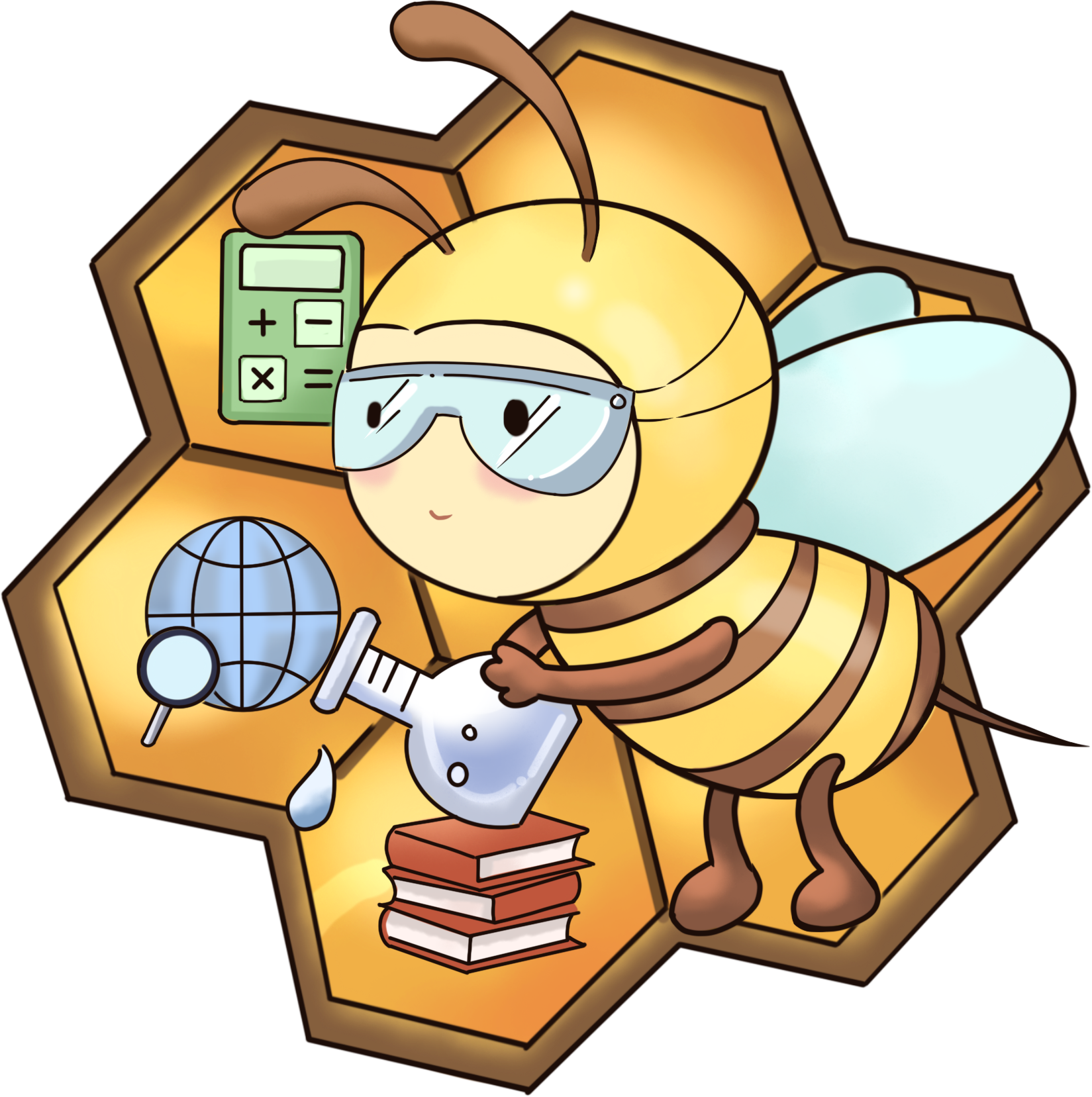}}
HoneyComb: A Flexible LLM-Based Agent System for Materials Science}

\author{
Huan Zhang\textsuperscript{1} , Yu Song\textsuperscript{1} , Ziyu Hou\textsuperscript{2} , Santiago Miret\textsuperscript{3}\thanks{Equal advising.} , Bang Liu\textsuperscript{1,4}\footnotemark[1]\thanks{Corresponding author.} \\
\textsuperscript{1}University of Montreal / Mila - Quebec AI, \textsuperscript{2}University of Waterloo, \\
\textsuperscript{3}Intel Labs, \textsuperscript{4}Canada CIFAR AI Chair \\
\{huan.zhang, yu.song, bang.liu\}@umontreal.ca \\
\{z26hou\}@uwaterloo.ca \\
\{santiago.miret\}@intel.com
}

\usepackage{times}
\usepackage{latexsym}
\usepackage{rotating}
\usepackage{flushend}

\usepackage{times}
\usepackage{graphicx}
\usepackage{latexsym}
\usepackage{multirow, makecell}
\usepackage{todonotes}
\usepackage{amsmath}
\usepackage{cleveref}
\usepackage{pifont}
\usepackage{pifont}
\usepackage{caption}
\usepackage{float}

\usepackage{microtype}
\usepackage{subfigure}
\usepackage{booktabs} 
\usepackage{xcolor}
\usepackage{setspace}
\usepackage{multicol}
\usepackage{threeparttable}
\usepackage{tabulary}
\usepackage{colortbl}
\usepackage{ragged2e}
\usepackage{amsmath,bm}
\usepackage{adjustbox}
\usepackage{listings}
\usepackage{xparse}
\usepackage{enumitem}
\usepackage{soul}
\usepackage{dirtytalk}
\usepackage{graphicx}

\usepackage{algorithm}
\usepackage{algorithmic}

\definecolor{codegreen}{rgb}{0,0.6,0}
\definecolor{cadmiumorange}{rgb}{0.93, 0.53, 0.18}

\definecolor{r1}{HTML}{EFCC00}
\definecolor{r2}{HTML}{EEE8AA}
\definecolor{r3}{HTML}{FFFFE0}
\definecolor{r4}{HTML}{dddcdc}
\definecolor{r5}{HTML}{c9d7f0}
\definecolor{r6}{HTML}{b2ccfb}
\definecolor{r7}{HTML}{9abbff}

\begin{document}
\maketitle
\begin{abstract}

The emergence of specialized large language models (LLMs) has shown promise in addressing complex tasks for materials science. Many LLMs, however, often struggle with distinct complexities of material science tasks, such as materials science computational tasks, and often rely heavily on outdated implicit knowledge, leading to inaccuracies and hallucinations. To address these challenges, we introduce \textit{HoneyComb}, the first LLM-based agent system specifically designed for materials science. HoneyComb leverages a novel, high-quality materials science knowledge base (MatSciKB) and a sophisticated tool hub (ToolHub) to enhance its reasoning and computational capabilities tailored to materials science. MatSciKB is a curated, structured knowledge collection based on reliable literature, while ToolHub employs an Inductive Tool Construction method to generate, decompose, and refine API tools for materials science. Additionally, HoneyComb leverages a retriever module that adaptively selects the appropriate knowledge source or tools for specific tasks, thereby ensuring accuracy and relevance. Our results demonstrate that HoneyComb significantly outperforms baseline models across various tasks in materials science, effectively bridging the gap between current LLM capabilities and the specialized needs of this domain. Furthermore, our adaptable framework can be easily extended to other scientific domains, highlighting its potential for broad applicability in advancing scientific research and applications.

\end{abstract}

\section{Introduction}
The emergence of large language models (LLMs) \citep{openai2024, claude3, touvron2023llama2, touvron2023llama1} in recent years has brought about the application of LLMs across a wide range of fields related to science and engineering \citep{ai4science2023impact}. This has resulted in a number of new benchmarks measuring the capabilities of language models to perform scientific tasks \citep{wang2023scibench, sun2024scieval, mirza2024large, song-etal-2023-matsci} along with the development of custom LLMs and LLM-based systems for scientific domains including chemistry \citep{bran2023chemcrow, boiko2023autonomous}, biology \citep{Madani_2023} and materials science \citep{song2023honeybee, gupta2022matscibert, walker2021impact}. 


While much progress has been made in adapting LLMs to common tasks in natural language processing \citep{song-etal-2023-matsci, song2023honeybee}, many more challenges remain in having LLMs be effective agents for real-world materials science tasks \citep{miret2024llms, miret2024perspective}. As highlighted by \citet{zaki2023mascqa}, LLMs often fail in performing important computational tasks for materials science. Common mistakes by most LLMs include conceptual errors where models fail to retrieve correct concepts, equations, or facts relevant to the questions, and factual hallucinations where incorrect information is generated. An analysis by \citet{miret2024llms} also revealed that LLMs by themselves struggle to generate relevant and correct information pertaining to specialized materials science tasks. While \citet{song2023honeybee} showed that instruction fine-tuning can help in improving performance, the high costs of continuous model training and fine-tuning make retraining-based approaches challenging to scale. This is further compounded by the fact that relevant knowledge is continuously updated through a diversity of knowledge sources, including pre-print servers (e.g., arXiv and ChemRxiv) , peer-reviewed literature, open encyclopedias (e.g. Wikipedia) and relevant websites. Furthermore, prior work has show that utilizing external tools may be a more promising approach to solve complex scientific tasks instead of relying entirely on an LLMs internal knowledge \citep{zheng2024step, buehler2024}. To jointly address these challenges, we propose transforming LLMs into LLM-based agents that access external knowledge and tools to boost their performance. This approach has already shown promise in adjacent domains, such as chemistry \citep{bran2023chemcrow, boiko2023autonomous} by enabling the the models to access real-time data and utilize computational as well as domain-specific tools. Altogether, the LLM-based agents showcase greater capabilities and performance compared to their native LLM counterparts.

In this paper, we present \textbf{HoneyComb}, the first, to the best of our knowledge, LLM-based agent system specifically designed for materials science. While there has been emerging research in LLMs for scientific domains, few studies have focused on developing comprehensive agent systems for materials science. Our work addresses two critical challenges: First, \textbf{MatSciKB} alleviates the challenge of obtaining reliable and relevant professional knowledge for materials. As such, MatSciKB ensures the agent has access to the most current and accurate information is essential for effective performance. Second, \textbf{Tool-Hub} provides materials science specific tools to augment the agent's capabilities. These tools enable the agent to perform specialized computational tasks and enhance its overall functionality. As detailed in \Cref{sec:experiments}, we observe that with the aid of MatSciKB and Tool-Hub, HoneyComb outperforms its native LLM counterparts in a more reliable manner given its ability to utilize up-to-date knowledge and tools.

\section{Background}

\subsection{LLMs for Material Science}

Advancements in text mining and information extraction from scientific publications have significantly benefited the application of LLMs for materials science \citep{kononova2021opportunities, swain2016chemdataextractor}. Early work include the development of specialized BERT models \citep{devlin2018bert}, such as MatSciBERT \citep{gupta2022matscibert} and MatBERT \citep{walker2021impact}. \citet{song2023honeybee} and \citet{xie2023darwin} leveraged instruction fine-tuning to develop a LlaMa-based \citep{touvron2023llama1} tailored to materials science that matched the capabilities of commercial LLMs at the time of publication. The emergence of powerful commercial LLMs \citep{openai2024, claude3} has further expanded the possibility of applying LLMs to materials science. Yet, commercial LLMs remain expensive, opaque in their methodology with consistent errors and shortcomings \citep{zaki2023mascqa, miret2024llms}, and open-source LLMs for materials science remain sparse. This motivates the need for a practical LLM-based system that is useful for real-world materials science tasks.


Given this need, we propose HoneyComb as an open-source system to augment the capabilities of diverse LLMs.  HoneyComb integrates specialized tools as well as a dynamic retrieval system to enhance the functionality any LLMs specifically for material science. By leveraging relevant knowledge source through MatSciKB and auxilliary tools through Tool-Hub, HoneyComb manages to improve the accuracy and relevance of the outputs of LLMs for materials science, while also addressing common challenges associated with static LLM applications in dynamic research fields.


\subsection{Tool-Based LLM Agents for Scientific Applications}

Prior work has shown success in expanding the capabilities of LLMs by augmenting their capabilities with diverse sets of tools \citep{qin2023toolllm, qin2023tool, chern2023factool, wang2024survey}. Many works rely on pre-built integration frameworks, such as LangChain \citep{topsakal2023creating}, to build the relevant interfaces between the LLMs and the desired capabilities, such as search engine APIs. \citet{wang2024survey} provides a recent survey of common approaches, challenges and applications of tool-based LLMs and their applications to various technology and scientific fields. 


One major application of tool-based LLMs is in query processing and optimization, where agents evaluate initial search results and iteratively refine queries to increase relevance and accuracy \cite{buehler2024, buehler2024mechgpt}. This approach addresses the limitations of isolated LLMs, which may struggle to handle ambiguous query contexts. In generating structured datasets for solar cell materials, agents gather pertinent information from a vast array of scientific papers to automate data input and synthesis \cite{xie2024, liu2024agent}. Furthermore, agents can utilize various tools to help answer specific questions by tapping into external resources \cite{cheng2024exploring}. For example, ChemCrow by \citet{bran2023chemcrow} integrates 18 expert-designed tools, such as literature search, molecule modification, and reaction execution, to autonomously execute chemical syntheses. Tool augmentation has also shown success in other research in the chemistry domain to enable real-world experiments using LLMs \citep{yoshikawa2023large, jablonka202314, boiko2023autonomous}. Coscientist by \citet{boiko2023autonomous}, for examples, relies on specialized tools to extend the capabilities of GPT4 and thereby invoke domain-specific functionalities that are not inherently present within the LLM alone. The success of agent-based approaches in adjacent domains motivates the creation of HoneyComb that extends the capabilities of LLMs specifically for materials science.

\section{HoneyComb}
\begin{figure*}[htbp]
  \centering 
  \includegraphics[width=\textwidth]{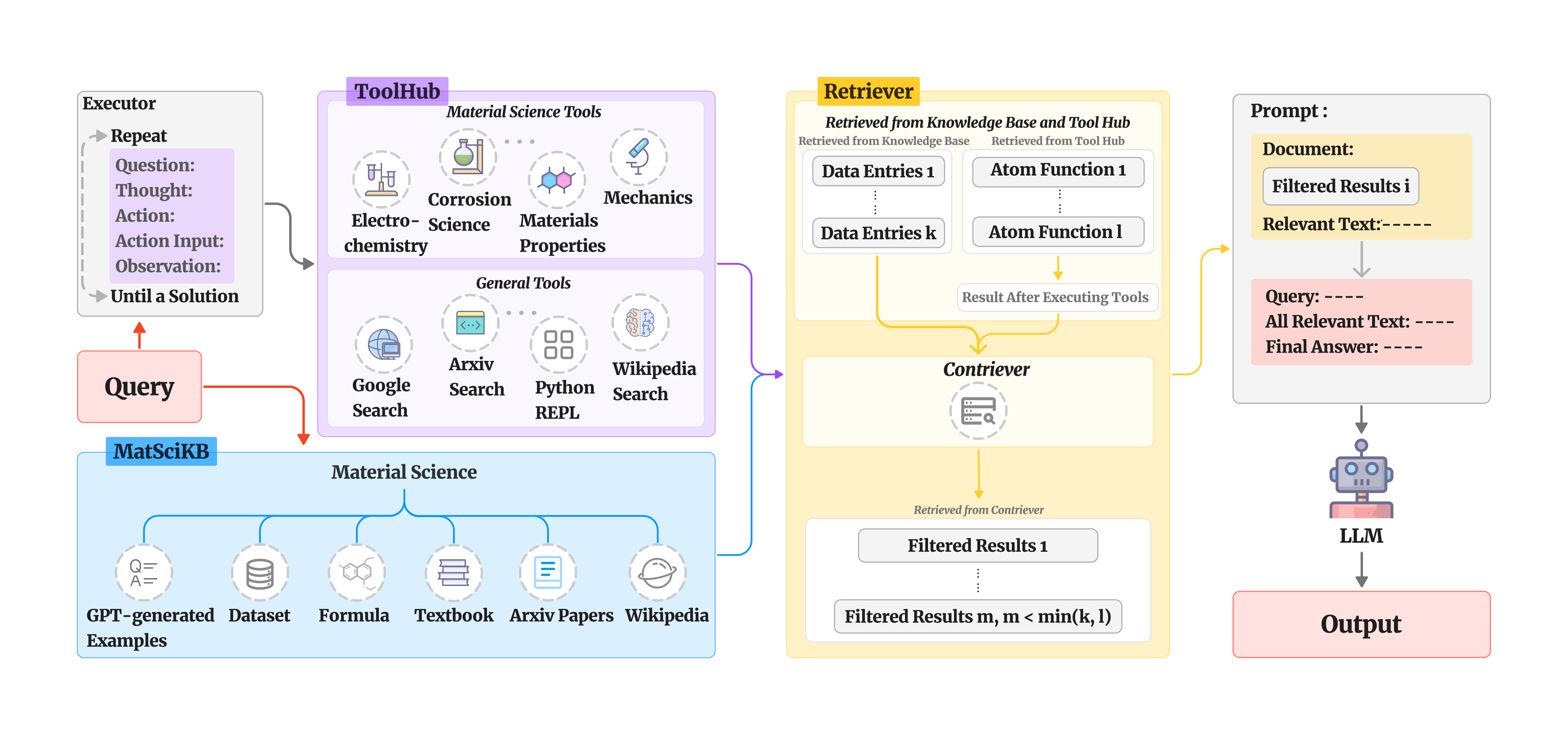}
  \caption{The overall architecture of HoneyComb. The model initiates with a query input that activates the knowledge retrieval phase, where pertinent data entries and atom function are extracted from the MatSciKB and Tool-Hub respectively. The Executor iterative calls the relevant tools from the Tool-Hub, evaluating and refining these calls until a solution that adequately solves the query emerges. The preliminary solution generated by these tools is combined with relevant data entries, and then undergoes further processing by the Retriever. Finally, the Retriever consolidates and filters these input, ultimately feeding them into the LLM for final answer generation.}
  \label{"model"}
\end{figure*}

In this work, we introduce HoneyComb, shown in Figure \ref{"model"}, a specialized agent system designed to advance materials science research. It integrates three key components: 1) \textit{MatSciKB}, a comprehensive knowledge base; 2) \textit{ToolHub}, which includes general tools for accessing up-to-date information broadly and specialized tools developed through an Inductive Tool Construction method for targeted material science queries; 3) \textit{Retriever}, utilizing a hybrid approach for efficient and precise information retrieval.

\subsection{MatSciKB} 

Our MatSciKB knowledge base integrates a diverse array of sources, as detailed in Table \ref{tab:matscikb statistics}. This collection is meticulously curated to include material science papers from ArXiv, relevant Wikipedia entries, textbooks, comprehensive datasets, pertinent mathematical formulas, and concrete GPT-generated examples tailored to material science. Each information source is thoroughly described in Appendix \ref{append:MatSciKB Construction}.

The architectural framework of MatSciKB is thoughtfully structured into 16 distinct categories pertinent to material science. These are detailed in Appendix \ref{Tree-Structure MatSciKB} and are organized in a tree-like structure. MatSciKB supports efficient searching and CRUD (Create, Read, Update, Delete) operations \cite{giannaros2023autonomous}, which are vital for both the application and ongoing maintenance of the database. Given the continuously evolving and expanding body of knowledge in the materials science domain, capabilities for efficient updates and searches based on real-time information are crucial for research and engineering applications. Additionally, our structured data approach enhances the integration of the diverse data sources commonly encountered in materials science \citep{miret2024llms}. This structure not only facilitates easy access and management but also allows for seamless extension to include additional data modalities.



\begin{table}[h]
\begin{tabular}{lr}
\hline
\multicolumn{2}{c}{\textbf{MatSciKB}}          \\ \hline
\multicolumn{1}{l}{\# Total Number of Data Entries}           & 38,469  \\
\multicolumn{1}{l}{\hspace*{0.5cm} \# Material Science Papers on Arxiv}           & 20,384  \\
\multicolumn{1}{l}{\hspace*{0.5cm} \# Wikipedia for Material Science}           & 3,620  \\
\multicolumn{1}{l}{\hspace*{0.5cm} \# Material Science Textbook}      & 1,930\\
\multicolumn{1}{l}{\hspace*{0.5cm} \# Material Science Dataset}      &  10,473 \\
\multicolumn{1}{l}{\hspace*{0.5cm} \# Material Science Formula}      &  57 \\
\multicolumn{1}{l}{\hspace*{0.5cm} \# GPT-generated Examples}   & 2,005  \\ 
\hline
\end{tabular}
\caption{Statistics of the MatSciKB knowledge base}
\label{tab:matscikb statistics}
\end{table}

\subsection{Tool-Hub}
The Tool-Hub in HoneyComb is bifurcated into \textit{General Tools} and \textit{Material Science Tools}. Both categories are organized through a unified interface that supports allow HoneyComb to make effective use of all available tools. \textbf{General Tools} provide researchers with access to the latest information filling gaps not covered by the static entries in MatSciKB. \textbf{Material Science Tools} are specifically designed to handle complex calculations and in-depth analyses. The details of the unified interface are further elucidated in Appendix \ref{unified interface}.

\textbf{General Tools Construction} 
\hspace{0.5em}

In materials science, one of the persistent challenges is keeping research outputs aligned with the diverse and ever-evolving data modalities that describe complex material systems \citep{miret2024llms}. The diversity of data sources and measurements leads to a rapid evolution of knowledge in this field, necessitating tools that can effectively access and integrate recent findings. Traditional static databases, while useful, often lag in capturing the newest research, creating gaps that can impede the currency and relevance of scientific analysis in real-time. Further, the need to efficiently process complex and dynamic computational tasks within the research workflow remains inadequately addressed, often requiring manual intervention which can introduce errors and inefficiencies. Thus, constructing tools that can handle varying data modalities and complexities, and that can adapt to the continual advancements in materials science, is essential for advancing the field.


To address these challenges, HoneyComb has been designed with innovative solutions that markedly enhance research capabilities in materials science. First, we integrated General Tools that provide direct access to current publications and facilitate dynamic discussions, as shown in Table \ref{HoneyComb-ToolHub}, effectively complementing the static MatSciKB. Secondly, recognizing the limitations of large language models (LLMs) in performing computational tasks, we implemented a Python REPL environment within HoneyComb. This environment is strategically utilized by the system when the agent, interacting with the Tool-Hub, identifies a need for basic numerical computations. The agent dynamically writes Python code for these tasks and executes it through the Python REPL, bypassing the LLM’s computational limitations. This automation not only streamlines data processing but also enhances the precision and reliability of numerical analyses in research activities.


\begin{table}[h!]
\small
\centering
\begin{tabular}{lc}
\midrule
\textbf{General Tools} & 6 \\
Google Search & \\
Google Scholar Search & \\
Arxiv Search & \\
Wikipedia Search & \\
YouTube Search & \\
Python REPL & \\
\bottomrule
\end{tabular}
\caption{ToolHub: General Tools}
\label{HoneyComb-ToolHub}
\end{table}

\textbf{\begin{algorithm}
\caption{Inductive Tool Construction}
\label{alg:ITC}
\begin{algorithmic}[1]
\REQUIRE Train Set \( D_{train} \), LLM \( M \)
\ENSURE Set of atom tools \( A \)
\STATE $A \leftarrow \emptyset$ \COMMENT{Initialize the set of atom functions}
\FOR{each question $q_i$ in $D_{train}$}
    \STATE $f_i \leftarrow M(q_i)$ \COMMENT{Generate specific function for $q_i$}
    \STATE Human verifies $f_i$
    \STATE Decompose $f_i$ into atom functions $a_i$
    \STATE $A \leftarrow A \cup a_i$ \COMMENT{Add atom functions to the set}
\ENDFOR
\RETURN $A$
\end{algorithmic}
\end{algorithm}
}

\textbf{Inductive Tool Construction for Materials Science}

Constructing domain-specific tool APIs presents significant challenges. It requires domain expert knowledge, and there are limited existing resources to draw upon. Additionally, many valuable data and tools are not open source, limiting their accessibility. Developing these tools is essential for effectively addressing the unique and complex queries inherent to materials science. The scarcity of pre-existing, specialized computational tools necessitates a methodical approach to tool construction and refinement.

We propose the \textit{Inductive Tool Construction} method, delineated in Algorithm \ref{alg:ITC} for domain-specific tool APIs construction. It adopts a systematic approach to fabricate and refine computational tools specifically designed for material science queries. The process initiates by selecting a random subset of computational questions from dataset \(D\), designated as \(D_{train}\) for training, with the residual questions forming \(D_{test}\). For each question \(q_i \in D_{train}\), a designated LLM, \(M\) (such as GPT-4), is tasked to generate a Python function \(f_i\) that addresses \(q_i\). After creation, each function \(f_i\) undergoes rigorous human verification to confirm its correctness. 

However, the above procedures cannot ensure the generalizability of the constructed tool APIs. Thus, in the post-validation stage, we further use \(M\) to decompose each \(f_i\) into fundamental, reusable components known as atomic function \(a_i\), which are crafted for extensive applicability across diverse queries, a detailed example is illustrated in Appendix \ref{Inductive Tool Construction Example}

\subsection{Agent-Tool Hub Interactions}

In HoneyComb, interactions between the agent and Tool-Hub are governed by a structured two-phase decision-making protocol. Our protocol emphasizes the critical selection and processing of data to ensure that only pertinent information influences the LLM's decisions. This approach is vital to prevent the degradation of model performance due to irrelevant or low-quality inputs \cite{llm_long_context_issue}.

1. Tool Assessor: During the initial phase, the Assessor evaluates both the incoming query and the extensive suite of tools within the Tool-Hub. This evaluation aims to identify a manageable subset of the most relevant tools that are best suited to address the specific requirements of the query. By filtering out irrelevant tools at this stage, we ensure that the Executor is provided only with pertinent information, thereby optimizing the model's focus and enhancing its capacity to solve the problem accurately.

\begin{figure}[htbp]
  \centering 
  \includegraphics[width=\columnwidth]{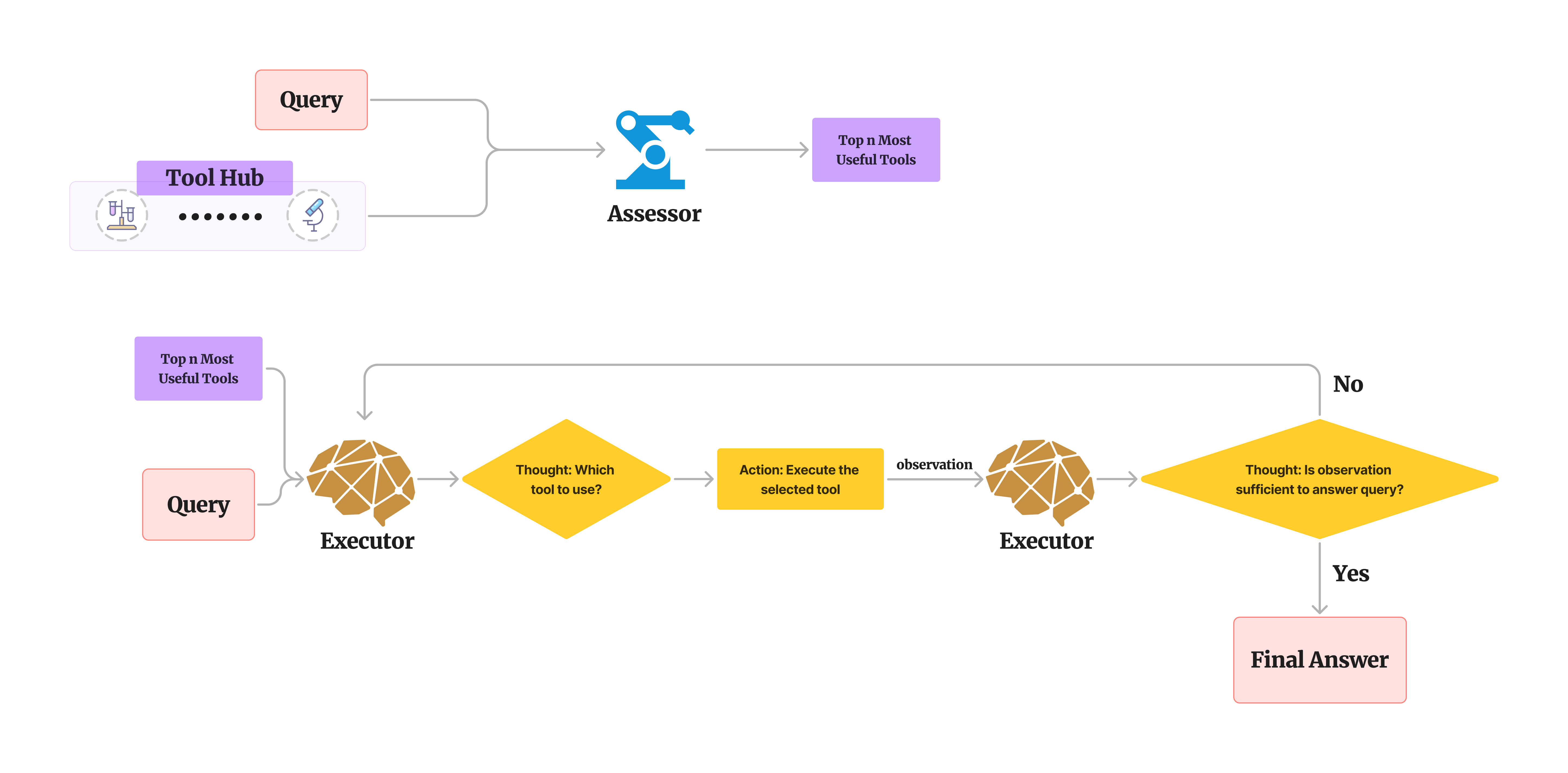}
  \caption{Tool Assessor and Executor interaction cycle in HoneyComb.}
  \label{fig:executor}
\end{figure}

2. Tool Executor: As illustrated in Figure \ref{fig:executor}, the Executor receives the original query along with the subset of tools selected by the Assessor. Upon evaluating the selected tools and query, the Executor engages in a \textit{thought} process to determine the most suitable tool for addressing the query. If the query's complexity exceeds the capacity of a single tool, the Executor recognizes the challenge and decomposes the query into smaller subquestions. The strategy allows for sequential tackling of each part, starting with the selection of the optimal tool for the initial subquestion. It then initiates the \textit{action} of executing the selected tool while inputting parameter values, termed \textit{action input}, derived from the query or subquestion. Upon execution, the tool generates a result termed \textit{observation}. Subsequently, the Executor engages in a reflective process to assess whether the observation adequately addresses the query. If the observation is adequate, it is finalized as the answer; if not, the process either reiterates with adjustments or progresses to the next subquestion if the original query was segmented into multiple parts.




\subsection{Retriever}

In this section, we present the retriever in HoneyComb which returns relevant texts or tools from MatSciKB and Tool-Hub when a specific contexts is given.
The retriever integrates both BM25 \citep{trotman2014improvements} and Contriever \citep{izacard2022unsupervised} model, leveraging their respective strengths to achieve optimal information retrieval performance. 

Specifically, the retriever employs a two-step strategy. 
Initially, BM25 utilizes efficient calculations of term frequency and inverse document frequency to rapidly process short text queries and keyword searches within long documents.
The primary advantage of BM25 lies in its computational simplicity and rapid response, allowing HoneyComb to extract the N most relevant knowledge points from an extensive materials science knowledge base, ensuring exceptional speed and efficiency. 
This approach enables the provision of basic relevance matching results in a minimal timeframe.


Subsequently, we employs a pre-trained deep learning models (i.e. Contriever) to generate embedding vectors and compute their similarity, facilitating the understanding of complex linguistic structures and semantic information. The strength of Contriever resides in its capability to comprehend and process intricate language structures, contextual information, and semantic relationships, thereby delivering more precise and comprehensive retrieval results. Although Contriever operates at a slower pace compared to BM25, it pulls the most relevant results from the knowledge base and memory, as well as from tools invoked through the Tool-Hub, extracting the top 3 results. Its ability to precisely handle complex queries and diverse documents ensures high accuracy and relevance. 

By combining BM25 and Contriever, our model adeptly responds to simple queries with speed while offering enhanced accuracy and relevance for complex queries. This hybrid approach ensures that the model is both efficient and capable of addressing sophisticated query requirements, thereby providing comprehensive, efficient, and precise information retrieval services.


\section{Experiments} \label{sec:experiments}



We conduct experiments on two question answering datasets, namely MaScQA~\cite{zaki2023mascqa} and SciQA~\cite{SciQ}, to investigating the ablility of HoneyComb in materials science tasks. 

MaScQA, derived from the Graduate Aptitude Test in Engineering (GATE) in India, is tailored to reflect the real-world complexity and variety of issues encountered in material science. This highly competitive examination assesses a comprehensive understanding of various undergraduate subjects \cite{gate2023overview, zaki2023mascqa}. With its 650 questions covering 14 domains such as thermodynamics, atomic structure, and mechanical behavior, the dataset showcases a wide range of question types, from Multiple Choice Questions (MCQs), Numerical Answer Type (NUM), and Matching Type (MATCH) to MCQs with numerical options (MCQN). Specifically designed for advanced problem-solving, this dataset is crucial for ensuring that our ToolHub functions effectively in actual material science research and applications. It demonstrates the HoneyComb framework's efficacy and adaptability in tackling complex material science issues within realistic scenarios. The second dataset, SciQA, comprises 11,679 multiple-choice questions that span the core disciplines of fundamental sciences from a variety of crowdsourced science exams \cite{SciQ}. This compilation not only underlines the dataset's comprehensive and interdisciplinary nature but also focuses on fostering a nuanced conceptual understanding. SciQA serves as a critical testbed to ascertain whether the HoneyComb framework can augment the LLM’s capabilities beyond its initial programming. By integrating supplementary information, it aids in addressing intricate queries and unraveling complex scientific concepts that may have been overlooked during the initial training phase of the LLM. By bridging real-world complexities with rigorous academic standards, these datasets ensure that our MatSciKB and ToolHub are not only versatile but also remain at the forefront of technological and scientific application.

The choice of models for our experiments was driven by the need to evaluate the HoneyComb framework's enhancement capabilities across a spectrum of large language models known for their robust performance in diverse applications. We selected GPT-3.5, GPT-4 \cite{openai2024}, LLaMA-2 \cite{touvron2023llama2}, and LLaMA-3 \cite{llama3modelcard} due to their widespread use and proven effectiveness in handling complex language tasks. These models, with LLaMA-2 and LLaMA-3 having parameter sizes of 7 billion and 8 billion respectively, represent the current state-of-the-art in generalized language understanding and provide a solid baseline for benchmarking. Additionally, we included HoneyBee\cite{song2023honeybee}, a specialized model with a parameter size of 7 billion, tailored specifically for materials science. The inclusion of both general-purpose and specialized models allows us to showcase how domain-specific adaptations through HoneyComb can elevate a model's functional scope beyond its original configuration, thus highlighting the adaptability and effectiveness of our framework.

    

\subsection{HoneyComb Evaluation}

\begin{table*}[ht!]
\begin{spacing}{1.15}
    \centering
    \vspace{0.5mm}
    \begin{adjustbox}{max width=\linewidth}
    \begin{threeparttable}
        \begin{tabular}{c|c|c|c|c|c|c|c|c|c|c}
            \toprule
            \bf{Dataset} & \bf{HoneyBee} & \bf{HoneyBee + \raisebox{-0.3em}{\includegraphics[scale=0.009]{figures/logo.png}}} & \bf{GPT-3.5} & \bf{GPT-3.5 + \raisebox{-0.3em}{\includegraphics[scale=0.009]{figures/logo.png}}} & \bf{GPT-4} &\bf{GPT-4 + \raisebox{-0.3em}{\includegraphics[scale=0.009]{figures/logo.png}}} & \bf{Llama2} &\bf{Llama2 + \raisebox{-0.3em}{\includegraphics[scale=0.009]{figures/logo.png}}} & \bf{Llama3} & \bf{Llama3 + \raisebox{-0.3em}{\includegraphics[scale=0.009]{figures/logo.png}}} \\
            
            \midrule
            MaScQA & 16.62 & 33.38 & 33.54 & 38.46 & 58.46 & 79.07 & 22.15& 36.31& 24.62& 47.23 \\
            \midrule
            SciQA & 33.96 & 79.69 & 90.69 & 90.83 & 90.84 & 96.54& 75.79& 78.66 & 93.00 & 93.32 \\
            \bottomrule
        \end{tabular}
    \end{threeparttable}
    \end{adjustbox}
\end{spacing}
\caption{HoneyComb evaluation with diverse LLMs including open-source LLMs (HoneyBee \citep{song2023honeybee}, LlaMa2 \citep{touvron2023llama2}, LlaMa3 \citep{llama3modelcard}) and commercial LLMs (GPT3.5, GPT4 \citep{openai2024}). The results show that HoneyComb consistently improves the performance of all LLMs for SciQA and MaScQA.}
\label{table_2}
\vspace{-1.5mm}
\end{table*}

\begin{figure*}[t!]
  \centering 
  \includegraphics[width=\textwidth]{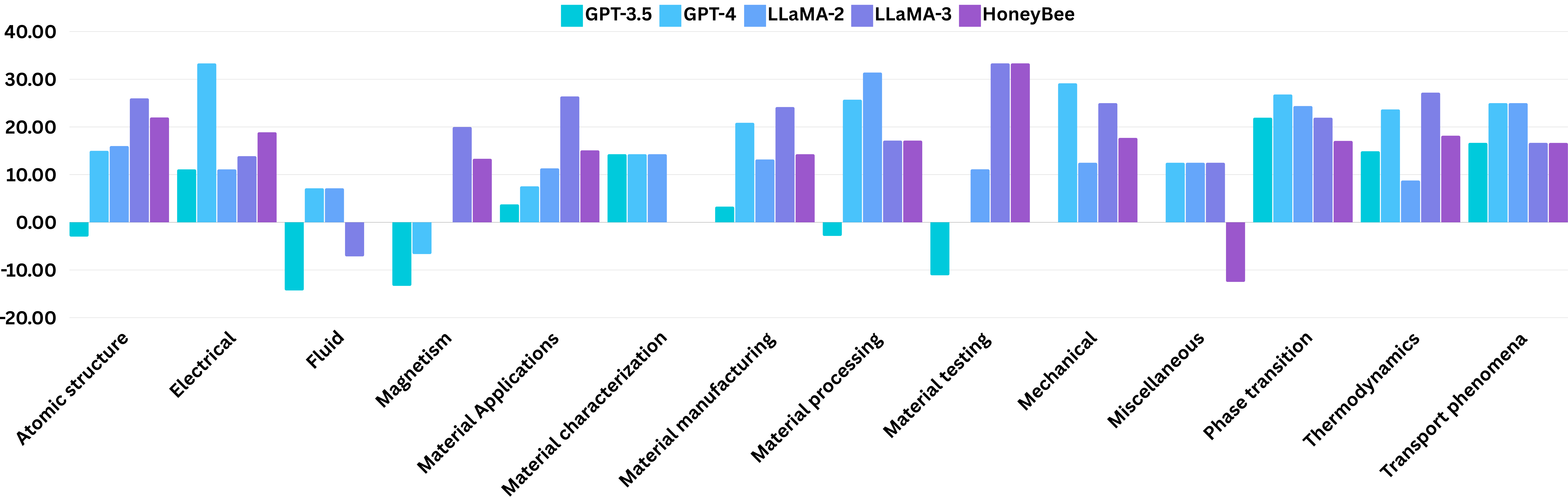}
  \caption{Improvements of various LLMs integrated with HoneyComb compared to relevant baseline LLMs for different materials science tasks. With few exceptions, HoneyComb improves the performance of all LLMs across all tasks showing the utility of tool augmentation.}
  \label{"topics"}
\end{figure*}

We evaluated the performance of various models on MaScQA and SciQA, including HoneyBee, GPT-3.5, GPT-4, Llama2, and Llama3, and demonstrated the effects of using the HoneyComb. The results are illustrated in Table \ref{table_2}

The experimental results show that all models based on HoneyComb achieved significant improvements in accuracy on both MaScQA and SciQA. Specifically, on the MaScQA dataset, models such as HoneyBee and GPT-4 experienced substantial improvements, with HoneyBee's accuracy improving by 16.76\% and GPT-4's by 20.61\%. Other models also showed notable enhancements, with improvements ranging from 4.92 to 14.16\%. On the SciQA dataset, the HoneyBee model saw a dramatic increase in performance, representing a huge improvement of 45.73\% . 
HoneyComb based on GPT-3.5 and Llama3 showed more modest enhancements of around 0.14\% to 0.32\% , whereas HoneyComb based on GPT-4 and Llama2 experienced considerable improvements of approximately 5.70\% and 2.87\% , respectively.

\subsection{HoneyComb Evaluation on MaScQA}


We assess the performance improvements when integrating the HoneyComb framework with various large language models across predefined topics within the MaScQA dataset, as shown in Figure \ref{"topics"}. The overall trend indicates that HoneyComb substantially enhances model performance. LLaMA-3 and HoneyBee exhibit impressive gains, particularly in 'Material Testing' where improvements of 33.34 percentage points are observed, showcasing HoneyComb's capability to effectively augment models with its advanced Tool-Hub and extensive MatSciKB.

However, GPT-3.5 displays a unique trend with declines across multiple topics including Atomic Structure, Fluid, Magnetism, Material Processing, and Material Testing. Despite having a higher baseline accuracy than LLaMA-3, LLaMA-2, and HoneyBee, GPT-3.5's performance dips more frequently when integrated with HoneyComb. This could be attributed to its training data's scope and depth, which, while extensive, may not align as effectively with HoneyComb's highly specialized material science enhancements. The sophisticated computational demands and the dynamic nature of materials science queries may expose limitations in GPT-3.5’s ability to adapt its pre-existing knowledge to the specific enhancements HoneyComb offers. This nuanced understanding highlights the importance of model and tool compatibility in achieving effective enhancements across diverse materials science domains, thereby informing further development and optimization of HoneyComb to ensure comprehensive and reliable support in all areas of materials science research.

\subsection{Ablation Study}
To study how each component of HoneyComb contributes to the overall performance, we conducted ablation studies in this section. 
We tested the performance of HoneyComb when retrieved only from MatSciKB or only from Tool Hub, respectively.
We also report results without retriever, in such situation there is no way for MatSciKB and ToolHub results to be fed into the model.
Experimental results are reported in Table \ref{table_ablation}.
\begin{table}[t!]
\caption{Ablation Study Results for MaScQA and SciQA based on GPT-4}
\label{table_ablation}
\begin{spacing}{1.0}
    \centering
    \vspace{0.5mm}
    \begin{adjustbox}{max width=\columnwidth}
    \begin{threeparttable}
    \small
        \begin{tabular}{ccccc}
            \toprule
            \bf{Benchmark} & \bf{MatSciKB} & \bf{ToolHub} & \bf{Retriever} & \bf{Accuracy} \\
            \midrule
            \multirow{4}{*}{MaScQA} 
            &  &  &  & 61.38 \\
            & & \checkmark & \checkmark & 73.23 \\
            & \checkmark & & \checkmark & 78.31 \\
            & \checkmark & \checkmark & \checkmark & 79.07 \\
            \midrule
            \multirow{4}{*}{SciQA}
            &  &  &  & 90.84 \\
            & & \checkmark & \checkmark & 96.34 \\
            & \checkmark & & \checkmark & 85.57 \\
            & \checkmark & \checkmark & \checkmark & 96.56 \\
            \bottomrule
        \end{tabular}
    \end{threeparttable}
    \end{adjustbox}
\end{spacing}
\vspace{-1.5mm}
\end{table}

The experimental results show that the best performance is achieved when both MatSciKB and ToolHub are used as reliable material knowledge references. 
HoneyComb improved the correctness on MaScQA and SciQA by 0.76\% and 10.99\% when compared to retrieving only from MatSciKB, and improved the correctness on MaScQA and SciQA by 5.84\% and 0.22\% when compared to retrieving only from Tool Hub. 
Therefore, we recommend that users retrieve HoneyComb from both sources together when deploying or using it.

\section{Conclusion}

In this work, we introduced HoneyComb, a pioneering LLM-based agent system tailored for materials science. HoneyComb integrates a meticulously curated materials science knowledge base (MatSciKB) and a dual-layered ToolHub of general and specialized computational tools. It combines three critical components: MatSciKB, an inductively constructed ToolHub, and a precision-focused Retriever module. This ensures HoneyComb provides accurate, up-to-date information and performs complex computational tasks reliably.

Experimental results show that HoneyComb outperforms contemporary general-purpose models (e.g. GPT and LLaMa series) and specialized models (e.g. HoneyBee) in materials science QA tasks. HoneyComb effectively bridges the gap between advanced large language models and the specific needs of materials science research, exemplifying how specialized agent systems can advance scientific research and serve as a blueprint for future advancements in other knowledge-intensive fields.
\section*{Limitations}

While HoneyComb significantly enhances the performance of current state-of-the-art models in various materials science QA tasks, there are limitations to its generali  zability and applicability beyond the specific datasets and tasks it was trained on. 
Materials science is a diverse and intricate field, and it remains unclear how well HoneyComb would perform on tasks outside the MaScQA and SciQA benchmarks, particularly for more complex and novel challenges in materials science. 
Such challenges may include designing synthesis recipes for new materials or predicting material properties.

Additionally, HoneyComb's reliance on high-quality LLMs for the knowledge base, tool construction, and retrieval processes can be a limitation. The performance of these components is contingent on the availability and capability of the underlying LLMs, which themselves may have inherent limitations. Furthermore, our work has primarily focused on the materials science domain, and further studies are required to evaluate how applicable and effective HoneyComb would be in other scientific fields.

\section*{Broader Impacts}

By expanding the HoneyComb agent system, HoneyComb has the potential to accelerate scientific discovery and innovation, contributing to a deeper understanding of complex materials systems. This could not only lead to advancements in materials design, development, and application but also promote the discovery and optimization of new materials, benefiting a wide range of industries. Additionally, the versatility and adaptability of HoneyComb enable it to tackle challenges across various scientific domains, further broadening its scope and impact.

Our research does not raise major ethical concerns.


\bibliography{custom}

\appendix
\section*{Appendix}

\section{MatSciKB Knowledge Source}
\label{append:MatSciKB Construction}

\begin{itemize}
    \item \textbf{ArXiv Paper}
        \begin{itemize}
            \item Included all papers indexed under the “material science” keyword on ArXiv.
            \item Data entries structured into key-value pairs: key is the paper title, and value is the abstract.
            \item \textbf{Data Entries Count:} 20,384
        \end{itemize}
    \item \textbf{Wikipedia Material Science Concepts}
        \begin{itemize}
            \item Scraped all 438 pages categorized under "Materials Science" on Wikipedia.
            \item Each section within a page was separated as a distinct data entry.
            \item Content formulated into key-value pairs, with keys as section titles and values as content.
            \item \textbf{Data Entries Count:} 3,620
        \end{itemize}
    \item \textbf{Material Science Textbook}
        \begin{itemize}
            \item Sourced 6 publicly available textbooks.
            \item Converted each textbook PDF file to text documents.
            \item Broke each textbook into data entries by each section in a chapter.
            \item Formulated data entries into key-value pairs, with keys as section titles and values as content.
            \item \textbf{Data Entries Count:} 1,930
        \end{itemize}
    \item \textbf{Material Science Dataset}
        \begin{itemize}
            \item Utilized the multiple-choice dataset SciQA.
            \item Extracted "support" column from the dataset that provides background knowledge for each question.
            \item Each extracted "support" is treated as a data entry, with keys as the knowledge piece and values as empty strings, emphasizing their concise and standalone nature.
            \item \textbf{Data Entries Count:} 10,473
        \end{itemize}
    \item \textbf{Material Science Formula}
        \begin{itemize}
            \item Formulas collected from Wikipedia's dedicated pages for material science formulas.
            \item Each formula is stored as key-value pair in the database, where the key represents the name of the formula and the value contains the formula equation itself.
            \item \textbf{Data Entries Count:} 57
        \end{itemize}
    \item \textbf{GPT-generated Examples}
        \begin{itemize}
            \item Used a specific prompt to generate 50 material science questions at a time, output in CSV format along with a confidence score. Please refer to Appendix \ref{gpt_prompt_genrated_examples} for the detailed prompt.
            \item Human reviewers then selected questions with higher confidence scores for inclusion in the dataset.
            \item Inspiration for question types was drawn from an external resource offering a wide range of material science questions and answers.
            \item The key-value pairs were structured with questions as the keys and answers as the values.
            \item \textbf{Data Entries Count:} 2,005
        \end{itemize}
\end{itemize}

\section{Prompt for GPT-Generated Examples}
\label{gpt_prompt_genrated_examples}

Please generate 50 instances of material science questions, specifically atomic structure and interatomic bonding, in a CSV format in the following order: question, answer, accuracy, confidence\_score
    - accuracy: for factual questions, please evaluate the answer by comparing it with known facts. this field should be a number between 0 and 1.
    - confidence\_score: how confident are you with the answer. this field should be a number between 0 and 1.
    - Here are sample instances without accuracy and confidence\_score: “In terms of which of the following properties, metals are better than ceramics?”,“ductility” "In the wave-mechanical model of an atom, what do degenerate energy levels have?","equal energy" "Which of the following molecules is diamagnetic?","CO"
  - Examples of generated instances:
    - "What is the valence electron configuration of carbon?","2s²2p²",0.95,0.85
    - "What type of crystal defect occurs when there is a line of irregularity in the lattice structure?","dislocation defect",0.96,0.91

\section{Tree-Structure MatSciKB}
\label{Tree-Structure MatSciKB}

MatSciKB is organized as a hierarchical tree with the parent node ``Material Science'' branching into 16 child nodes representing specific domains within materials science. Below is a simplified representation of this structure:

\texttt{\{\\
  "Material Science": \{\\
    \quad"Children": \{\\
      \qquad"Thermodynamics": \{"Children": \{"KB\_1": \{\}, "KB\_2": \{\}, "KB\_3": \{\}\}\},\\
      \qquad "Atomic Structure": \{"Children": \{"KB\_4": \{\}, "KB\_5": \{\}, "KB\_6": \{\}\}\},\\
      \qquad ...\\
      \qquad "Miscellaneous": \{"Children": \{"KB\_7": \{\}, "KB\_n": \{\}, "KB\_n+1": \{\}\}\}\\
    \}
  \}
\}
}

Each child node encompasses knowledge base (KB) data entries relevant to its category. In the construction of MatSciKB, we predefined 16 topics that align with core areas in materials science. They are 'Miscellaneous', 'Material testing', 'Fluid', 'Material characterization', 'Magnetism', 'Transport phenomena', 'Material processing', 'Electrical', 'Phase transition', 'Material Applications', 'Material manufacturing', 'Mechanical','Atomic structure', 'Thermodynamics', "Formula", "Fundamental\_Science\_Knowledge"]

To categorize the data entries within these nodes, we utilized BertTopic, a state-of-the-art topic modeling tool based on transformers and c-TF-IDF, which automatically identifies and clusters documents with high granularity and contextual relevance \cite{vaswani2023attention, grootendorst2022bertopic}. The integration of BertTopic allowed for the dynamic clustering of MatSciKB entries into 16 predetermined categories.

The process involved the following steps:

\begin{enumerate}
  \item \textbf{Initial Clustering}: BertTopic was applied to cluster all data entries into more than the target number of categories, based on the textual content of each entry.
  \item \textbf{Cluster Analysis and Selection}: Human reviewers analyzed each cluster, identifying those whose common keywords and themes closely aligned with one of the predefined 16 topics.
  \item \textbf{Category Assignment}: Entries from clusters that aligned well with a predefined topic were assigned to that category, and then removed from the dataset.
  \item \textbf{Iterative Refinement}: The remaining entries underwent subsequent rounds of clustering and analysis. This process was repeated until no entries were left unclassified.
\end{enumerate}

\section{Tools Unified Interface Using LangChain}
\label{unified interface}
LangChain is an advanced framework designed to enhance applications that utilize LLM by offering standardized interfaces for various modules \cite{langchain_github}. This framework facilities the seamless integration and efficient management of LLM with external tools and systems. Utilizing LangChain, HoneyComb has developed a unified interface that standardizes the integration of a wide array of tools.

In HoneyComb, the unified interface provided by LangChain ensures that all tools, regardless of their specific function, are treated as standardized LangChain objects. This standardization is achieved by defining each tool with a consistent set of attributes:

\begin{enumerate}
    \item \textbf{Function Signature}: Each tool is defined with a clear function signature that specifies input and output types,
    \item \textbf{Metadata Description}: Each tool is accompanied by metadata that describes its purpose, suitable use cases, parameters description.
\end{enumerate}
\begin{flushleft}
Examples of function signatures and metadata descriptions in HoneyComb are:
\begin{itemize}
    \item \textbf{Google Search}
            \begin{itemize}
\item \textbf{Function Signature}: Google\_Search(query: str, timeout: Optional[int] = 30) -> str
\item \textbf{Metadata Description}: General web search for up-to-date information across various topics.
            \end{itemize}
    \item \textbf{Wikipedia Search}
        \begin{itemize}
            \item \textbf{Function Signature}: Wikipedia\_Search(topic: str, summarize: bool = True) -> str
            \item \textbf{Metadata Description}: Retrieves and optionally summarizes detailed Wikipedia articles, particularly useful for quick reference checks.
        \end{itemize}
    \item \textbf{A Sample Mass Flow Rate Tool}
        \begin{itemize}
            \item \textbf{Function Signature}: calculate\_initial\_mass\_flow\_rate(args: str) -> float
            \item \textbf{Metadata Description}: See figure \ref{fig:metadata description of mass flow rate}.
    \end{itemize}
             \begin{figure}[htbp]
              \centering 
              \includegraphics[width=\columnwidth]{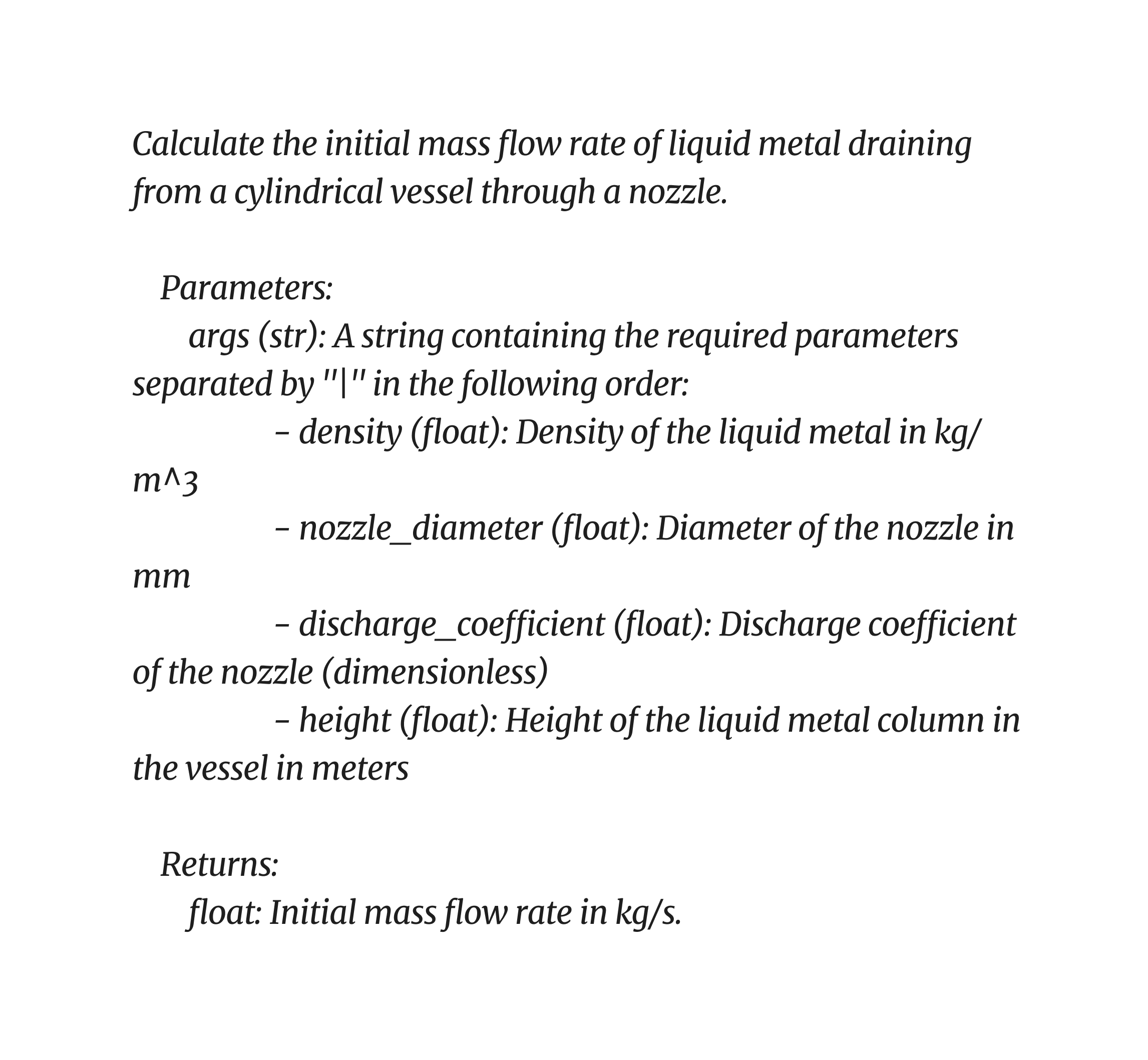}
              \caption{Metadata Description of a Sample Mass Flow Rate Tool}
              \label{fig:metadata description of mass flow rate}
            \end{figure}

\end{itemize}
\end{flushleft}

\section{Examples of Inductive Tool Construction}
\label{Inductive Tool Construction Example}
\begin{figure*}[htbp]
  \centering 
  \includegraphics[width=\textwidth]
  {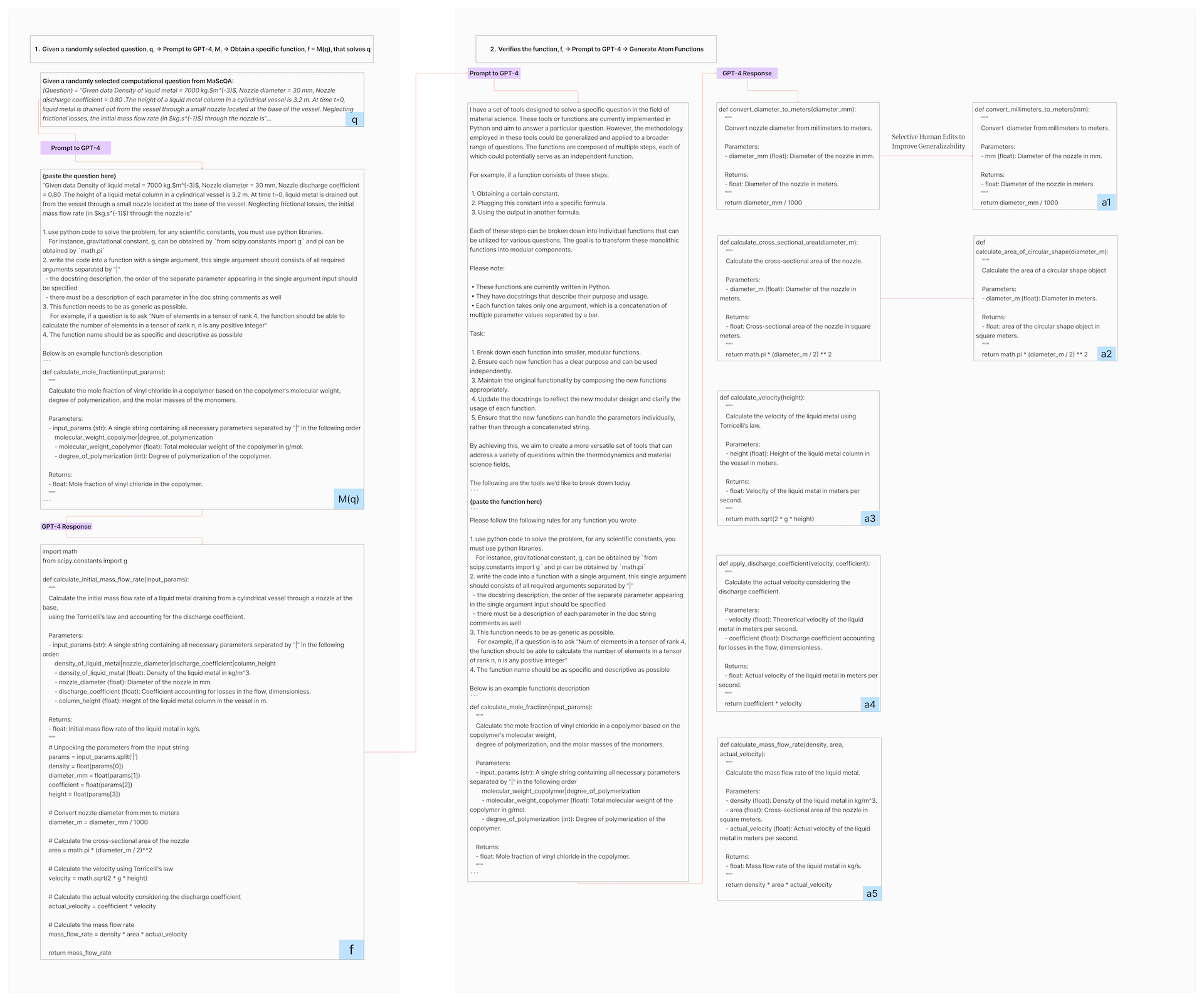}
  \caption{An example of inductive tool construction}
  \label{fig:inductive tool construction}
\end{figure*}

See figure \ref{fig:inductive tool construction} for a detailed example illustrating how inductive tool construction work.

\end{document}